\pdfoutput=1
\documentclass[11pt]{article}

\usepackage[final]{acl}

\usepackage{times}
\usepackage{latexsym}
\usepackage{booktabs}
\usepackage{amssymb}
\usepackage{pifont}
\usepackage[T1]{fontenc}
\usepackage[utf8]{inputenc}

\usepackage{microtype}

\usepackage{inconsolata}

\usepackage{graphicx}

\title{Exploring the Inquiry-Diagnosis Relationship with Advanced Patient Simulators}

\author{Zhaocheng Liu, Quan Tu\textsuperscript{\textdagger}, Wen Ye, Yu Xiao, Zhishou Zhang, \\ \textbf{Hengfu Cui}, \textbf{Yalun Zhu}, \textbf{Qiang Ju}, \textbf{Shizheng Li}, \textbf{Jian Xie}
   \\ 
Baichuan Inc.\\ \textsuperscript{\textdagger} Gaoling School of Artificial Intelligence, Renmin University of China
\\
\texttt{lio.h.zen@gmail.com,} \texttt{quantu@ruc.edu.cn,} \\ \texttt{\{yewen, xiaoyu, zhangzhishou, cuihengfu\}@baichuan-inc.com} \\ \texttt{\{zhuyalun, liulifeng, lishizheng, richard\}@baichuan-inc.com}
\\
  \small{
    \textbf{Correspondence:} \href{mailto:liulifeng@baichuan-inc.com}{liulifeng@baichuan-inc.com}
  }
}

\begin{document}
\maketitle
\begin{abstract}

Recently, large language models have shown great potential to transform online medical consultation. Despite this, most research targets improving diagnostic accuracy with ample information, often overlooking the inquiry phase. Some studies try to evaluate or refine doctor models by using prompt-engineered patient agents. However, prompt engineering alone falls short in accurately simulating real patients. 
% These challenges highlight the need for a new paradigm for patient simulation. 
We need to explore new paradigms for patient simulation. Furthermore, the relationship between inquiry and diagnosis remains unexplored. This paper extracts dialogue strategies from real doctor-patient conversations to guide the training of a patient simulator. 
% With actual patient profiles, o
Our simulator shows higher anthropomorphism and lower hallucination rates, using dynamic dialogue strategies. This innovation offers a more accurate evaluation of diagnostic models and generates realistic synthetic data. We conduct extensive experiments on the relationship between inquiry and diagnosis, showing they adhere to Liebig's law: poor inquiry limits diagnosis effectiveness, regardless of diagnostic skill, and vice versa. The experiments also reveal substantial differences in inquiry performance among models. To delve into this phenomenon, the inquiry process is categorized into four distinct types. Analyzing the distribution of inquiries across these types helps explain the performance differences. The weights of our patient simulator are available \textcolor{blue}{\href{https://github.com/PatientSimulator/PatientSimulator}{here}}.
\end{abstract}

\section{Introduction}

\begin{figure*}[t]
    \centering
    \includegraphics[width=1\linewidth]{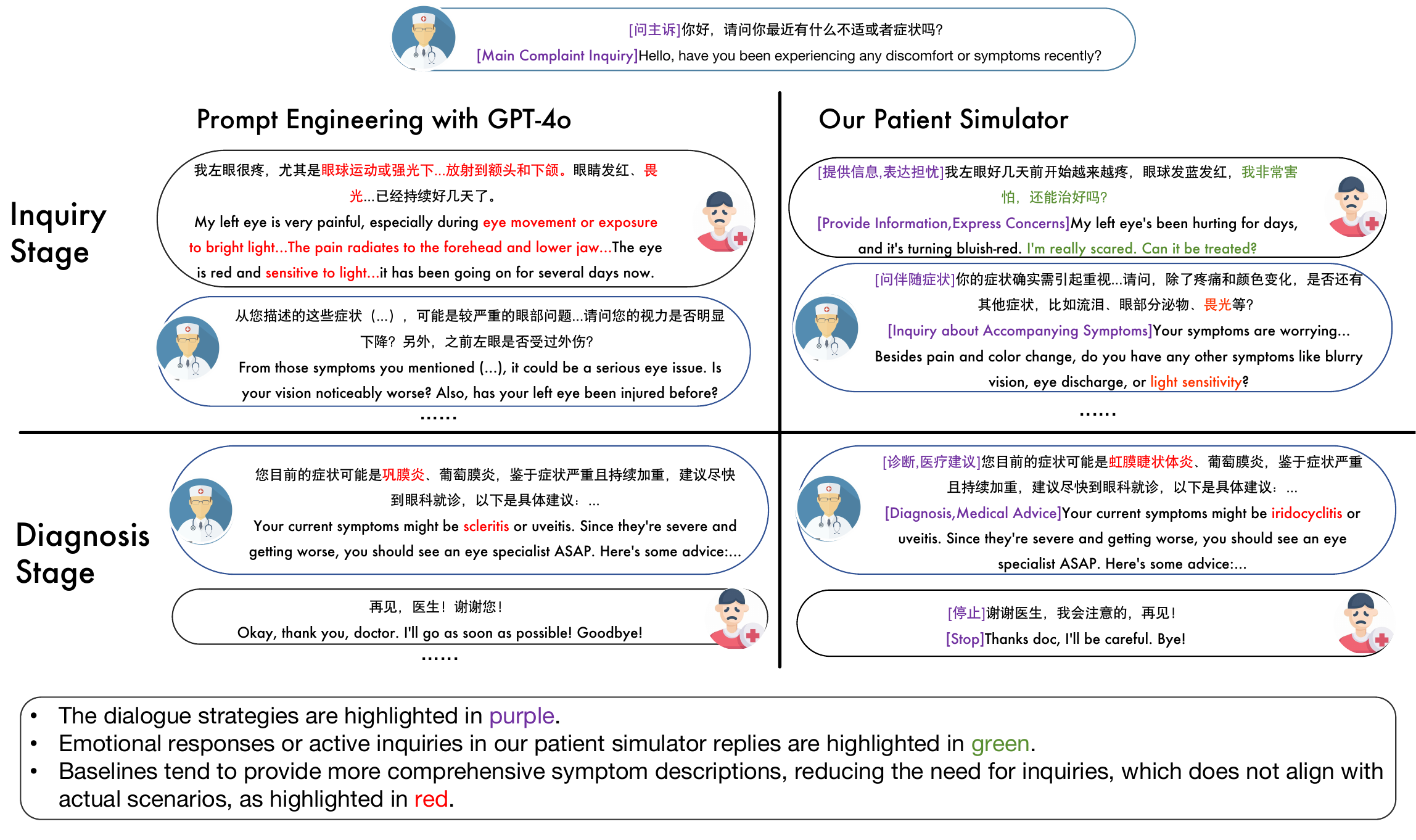}
    \caption{Our patient simulator (right) is compared to the baseline simulator (prompt engineering with GPT-4o, left) using identical patient records and doctor model.}% \textcolor{red}{drop this?}Online consultation dialogues are divided into inquiry and diagnosis stages. Based on the predefined set of dialogue strategies outlined in this paper, the dialogue strategies output by our model are highlighted in \textcolor[rgb]{0.4392,0.1882,0.6274}{purple}. The output from our patient simulator may contain emotions or proactive questions, marked in \textcolor[rgb]{0.345,0.55686,0.192}{green}. Baselines tend to provide more comprehensive symptom descriptions, reducing the need for inquiries, which does not align with actual scenarios, as highlighted in \textcolor[rgb]{1.0,0.,0.}{red}.}
    \label{fig:fig1}
\end{figure*}
Online medical consultation (OMC) \cite{al2015online,kessler2023online}, emerges as a revolutionary medical service that significantly improves the accessibility of healthcare, particularly in areas lacking adequate medical resources. However, compared to traditional face-to-face consultations, online consultations present notable limitations. The absence of direct physical examinations and auxiliary diagnostic tools requires physicians to depend entirely on inquiries to collect necessary information. This reliance limits a comprehensive evaluation of the patient’s health condition and substantially complicates the diagnostic process.

In recent years, large language models (LLMs) have demonstrated remarkable capabilities across various domains and tasks.
Notably, models such as OpenAI’s o1 \cite{openaio1} have introduced groundbreaking reasoning abilities by employing techniques akin to an internalized chain-of-thought \cite{wei2022chain} process.
% todo cites
Building on the core strengths of general-purpose LLMs, domain-specific models \cite{tian2023chimed,saab2024capabilities,chen2024huatuogpt,zhang2024ultramedical,singhal2025toward} tailored for healthcare have also emerged.
In the field of clinical medicine, numerous studies \cite{jin2021disease,jin2019pubmedqa,xie2024preliminary,tu2024towards,schmidgall2024agentclinic,liu2024medchain} have validated the performance of these models, suggesting their potential for transformative applications in medical practice.
For instance, on the MedQA (USMLE) benchmark \cite{jin2021disease}, models like GPT-4 \cite{achiam2023gpt} with MedPrompt \cite{nori2023can}, Med-Gemini-L 1.0 \cite{saab2024capabilities}, and o1-preview \cite{xie2024preliminary} have achieved performance levels surpassing those of human experts.

However, most doctor models focus on improving diagnostic accuracy under relatively sufficient information conditions, diverging from the challenges of online consultations.
% which clearly diverges from the primary challenges faced in online consultations.
OMC can be divided into two key stages: inquiry and diagnosis. 
Existing research has largely overlooked the inquiry stage, limiting understanding of its relationship with diagnosis.
% Existing research has paid relatively little attention to the inquiry stage, and this oversight have hindered a deeper understanding of the relationship between inquiry and diagnosis.
Some studies \cite{li2024mediq,tu2024towards,schmidgall2024agentclinic,qiu2024interactive,li2024agent} simulate clinical environments via prompt-engineered patient agents, yet these agents fail to replicate real patient behaviors.
% Some studies \cite{li2024mediq,tu2024towards,schmidgall2024agentclinic,qiu2024interactive,li2024agent} have attempted to evaluate or improve doctor models by simulating clinical environments.
% These studies use prompt engineering to construct patient agents, but the simulated results show significant discrepancies compared to the behavior of real patients.
In real-life scenarios, patients often show concern and anxiety about their condition in their responses.
When describing initial symptoms, they urgently highlight the issues they are most worried about, rather than giving a comprehensive list of all symptoms.
Real patients also tend to ask questions actively to ease their emotions.
They may disengage if doctors repeat questions.
% Moreover, real-life patients cannot always patiently answer questions.
% If a doctor (especially a doctor agent) keeps asking questions repeatedly, real patients may decide to exit the conversation or refuse to respond.
These challenges are difficult to address solely with prompt engineering, making it necessary to explore a new paradigm for simulating patients. 
Furthermore, current studies offering dynamic simulation environments have yet to thoroughly explore the relationship between inquiry and diagnosis. % and how it affects overall outcomes.

In this paper, we extract patient dialogue strategies from real doctor-patient conversations to guide the development of a patient simulator.
Initially, we annotate and standardize open-source real doctor-patient conversations using LLMs, and then summarize a set of patient dialogue strategies.
We manually select strategies that meet specific criteria, such as ensuring dialogue rounds are complete and excluding follow-up visits in favor of initial consultations.
% Given that the amount of usable training data is limited after selection and lacks corresponding medical records, we synthesize doctor-patient dialogue data using in-context learning.
Due to limited usable training data and the absence of medical records, we synthesize doctor-patient dialogues through in-context learning.
This synthesis relies on two types of inputs: (1) patient records (similar to context in MedQA), and (2) our curated dialogue strategy set.
We train our model entirely on the synthesized dialogues with corresponding patient records.
% Ultimately, we train our model entirely on the synthesized doctor-patient dialogue data along with the relevant medical records.
% After evaluation, our patient simulator demonstrates a lower hallucination rate in terms of dialogue and medical record consistency, although the rate of unrelated responses is slightly higher.
% In addition, there is a significant improvement in anthropomorphism, including emotions and dialogue strategies.
% It is important to note that the slightly higher rate of unrelated responses does not necessarily indicate a worse model performance because real patients also exhibit some degree of refusal to answer.
% Our unrelated responses mainly occur towards the end of dialogues, especially when the doctor model poses numerous questions, prompting the patient simulator to ask questions actively instead of responding directly.
After a thorough evaluation, our patient simulator shows a notable reduction in hallucination rates, achieving 0.31\% compared to the state-of-the-art (SOTA) rate of 3.71\%.
And there is a significant improvement in anthropomorphism, with our simulator scoring 0.87, outperforming the strongest baseline of 0.31.
Although our simulator exhibits a higher rate of unrelated responses at 4.79\%, compared to the best baseline of 0.93\%, this does not necessarily indicate inferior model performance.
This is attributed to real patients often exhibiting some refusal to answer, particularly in the latter part of inquiries. 
Our patient simulator also reflects this behavior.

Based on our patient simulator, we conduct extensive experiments to explore the relationship between inquiry and diagnosis. % and their impact on the accuracy of final diagnoses.
Utilizing our patient simulator to fix patient simulations, while interacting with different doctor models for a fixed number of rounds to generate inquiry records.
Each inquiry record is diagnosed using various doctor models.
Upon analyzing the diagnostic accuracy of the inquiries produced by different doctor models, we find that some models consistently yield inquiries with significantly high or low accuracy, regardless of which doctor model performed the diagnosis.
This indicates that \textbf{there are significant differences in the inquiries generated by different doctor models}.
Comparing high-quality and low-quality inquiries and observing the accuracy differences after diagnosis by doctor models with varying diagnostic capabilities, we propose that \textbf{the inquiry-diagnosis relationship follows Liebig's law}. 
When inquiry quality is inadequate, even strong diagnostic capabilities are insufficient for achieving good outcomes, and vice versa.

To further analyze the differences in inquiry processes among different doctor models, we categorize the inquiries into four types: (1) chief complaint inquiry; (2) specification of known symptoms; (3) inquiry about accompanying symptoms; (4) gathering family or medical history.
We calculate the distribution of inquiry records across these four types for different inquiry models.
By comparing the distribution differences and diagnostic accuracy, we uncover a certain correlation.
For instance, when a model asks more questions to specify known symptoms, resulting in relatively fewer inquiries of other types, the final diagnostic accuracy tends to be lower. 
Our findings suggest that optimizing the allocation of inquiries within typically 3 to 5 rounds, which patients can comfortably accept, presents a valuable research problem.

In summary, our contributions are as follows:
\begin{itemize}
    \item We propose a novel patient simulator guided by real dialogue strategies that effectively addresses the limitations of prompt-engineered patient agents by demonstrating higher anthropomorphism, reduced hallucination rates, and dynamic dialogue behavior.
    \item Extensive experiments explore the relationship between inquiry and diagnosis, showing that they align with Liebig's law. Poor inquiry quality constrains diagnostic accuracy, regardless of diagnostic capability, and vice versa.
    \item The inquiry process is systematically categorized into four distinct types to analyze discrepancies in model performance, thereby providing insights into the significant differences in inquiry quality across models.
\end{itemize}

\section{Patient Simulator}

\begin{figure}
    \centering
    \includegraphics[width=0.9\linewidth]{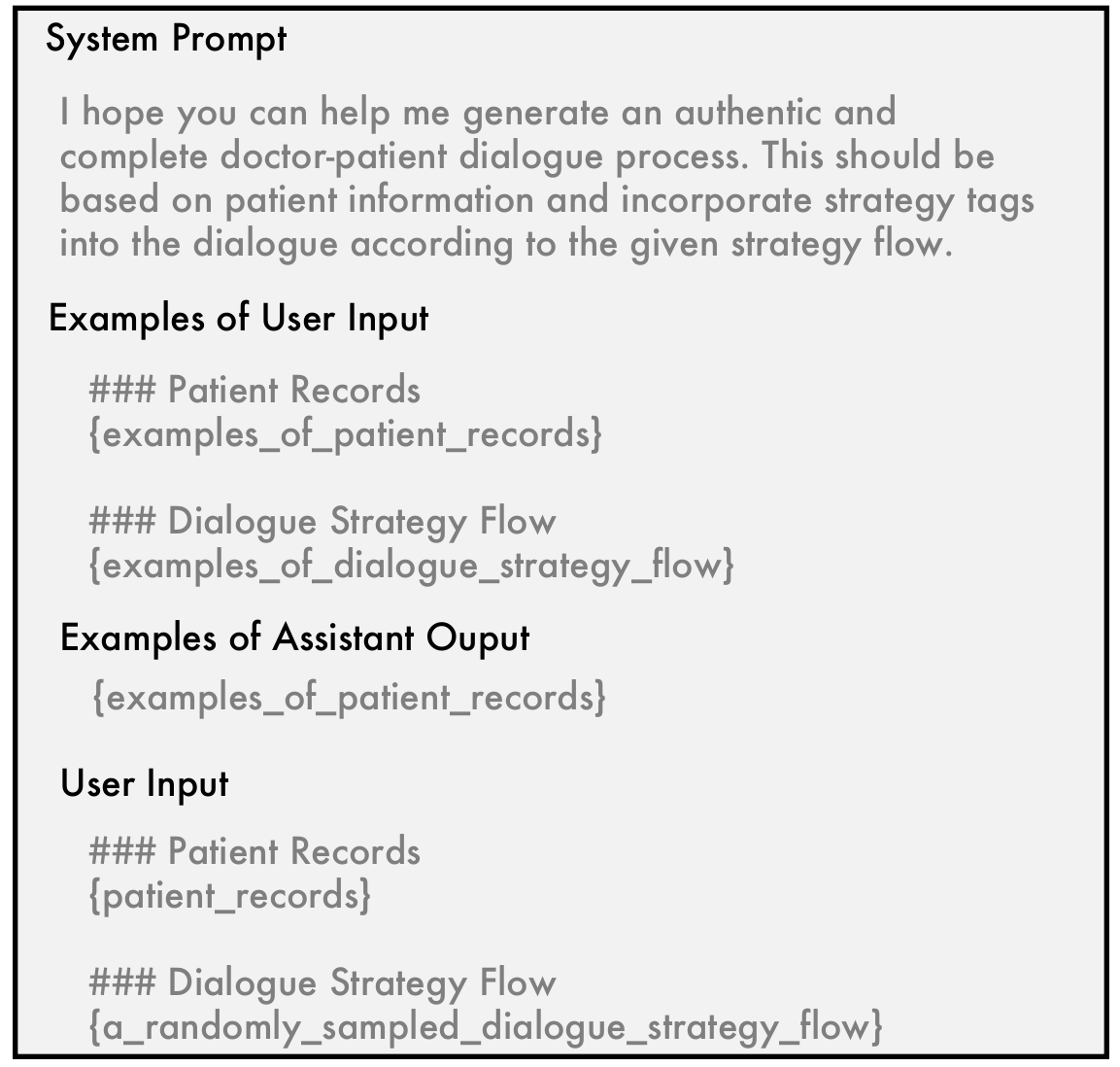}
    \caption{Prompts for synthesizing patient simulator training dialogues.}
    \label{fig:prompts_for_synthesizing_dialogues}
\end{figure}
% This section outlines the training methodologies and evaluation results of our patient simulator.
\subsection{Methods}

Some studies \cite{li2024mediq,tu2024towards,schmidgall2024agentclinic,qiu2024interactive,li2024agent} have attempted to assess or enhance doctor models by creating simulated clinical environments.
In these studies, prompt engineering is often used to construct patient agents.
% However, the interactions produced by this method significantly differ from those with real patients.
% Specifically, real-life patients may exhibit concerns and anxieties about their medical conditions when they communicate.
% They tend to urgently express the primary symptom that worries them the most during the initial description, rather than providing an exhaustive list of symptoms.
% Additionally, real patients may actively seek information to allay their emotional distress.
% Moreover, they are unlikely to remain patient indefinitely when responding to questions.
% If physicians (especially doctor agents) persistently ask questions, real patients might choose to terminate the conversation or refuse to answer.
Unfortunately, real patient behaviors are difficult to replicate through prompt engineering alone.

In order to simulate real patients as accurately as possible, it is necessary to rely on authentic doctor-patient dialogue datasets.
In this paper, we utilize the MedDialog \cite{zeng2020meddialog} dataset which is distributed under CC BY-NC 4.0\footnote{\url{https://creativecommons.org/licenses/by-nc/4.0/}}.
We sample the data to ensure it is thoroughly anonymized and free from any personal identifiers or offensive content.
Initially, we conduct essential data screening to remove non-consultative records (e.g., patient scheduling and registration) and to select complete initial consultation dialogues.
Then, we manually provide a seed set of commonly used dialogue strategy tags found in doctor-patient interactions.
GPT-4o \cite{openai4o} is employed to expand this seed set, resulting in a candidate set of dialogue strategy tags (see Appendix \ref{candidate_set_of_dialogue_strategy_tags}).
Based on the candidate set of dialogue strategy tags, GPT-4o is further used to annotate the selected complete initial consultation dialogues.
Each dialogue's tags are concatenated in sequence to form a dialogue strategy flow.
Finally, high-quality dialogue strategies are manually selected from the deduplicated set. Examples of selected dialogue strategy flows can be found in Appendix \ref{examples_strategy_appendix}.
% For example, the following is a selected dialogue strategy flow: 
% \textbf{[Doctor: Greeting],[Patient: Greeting],[Doctor: Chief Complaint Inquiry],[Patient: Provide Information],[Patient: Express Concerns],[Doctor: Gathering Family or Medical History], [Patient: Provide Information], [Doctor: Evaluation], [Doctor: Explanation], [Patient: Explanation Request], [Doctor: Answering], [Patient: Seek Advice], [Doctor: Medical Advice], [Patient: Discuss Treatment Options], [Doctor: Arrangement], [Patient: Seek Help], [Doctor: Medical Advice], [Patient: Thanks], [Doctor: Goodbye], [Patient: Stop]}.

Due to the limited availability of usable patient-doctor dialogue data for training after selection, and the absence of corresponding medical records, we synthesize patient-doctor dialogue data to facilitate the training process.
We utilize the Chinese medical record dataset released by CCKS 2019 \cite{han2020overview} as a candidate set of medical records.
We sample the data to ensure it is thoroughly anonymized and free from any personal identifiers or offensive content.
In each data synthesis iteration, a medical record is randomly selected, and a dialogue strategy flow is randomly chosen from the curated set of dialogue strategy flows.
Through in-context learning, we synthesize patient-doctor dialogues that align with the selected dialogue strategy flow. 
For detailed prompts, please refer to Figure \ref{fig:prompts_for_synthesizing_dialogues}.
Since the medical records are in Chinese, we synthesize only Chinese dialogues, limiting the simulator's performance in English scenarios.

% \begin{figure}
%     \centering
%     \includegraphics[width=\linewidth]{assets/patient_infer.pdf}
%     \caption{The system prompt of our patient simulator.}
%     \label{fig:patient_sys}
% \end{figure}
The format of this synthetic doctor-patient dialogue is shown on the right side of Figure \ref{fig:fig1}.
Each round of conversation between the doctor and the patient is preceded by several dialogue strategy tags.
We construct a supervised fine-tuning (SFT) dataset entirely based on this synthetic doctor-patient dialogue dataset.
Specifically, in the training and prediction phases, our patient simulator requires only the input of patient medical records into a simple system prompt (see Appendix \ref{patient_sys_appendix}).
Given a doctor-patient dialogue $\{d_1, p_1, d_2, p_2, \ldots, d_n, p_n\}$, where $d_i$ represents the i-th round of doctor dialogue and $p_i$ represents the i-th round of patient dialogue.
We divide it into $n$ SFT data instances, that is, $$\{d_1, p_1\}, \{d_1, p_1, d_2, p_2\}, ..., \{d_1, p_1, \ldots, d_n, p_n\}.$$
It is important to note that for each SFT data instance, we only retain the dialogue strategy tags for the label (the last turn of the patient dialogue).
The strategy tags in the preceding dialogues are removed.
This is to align with the estimated scenarios of the patient simulator, as the doctor model is not expected to output our dialogue strategy tags.
The model needs to learn to predict the appropriate dialogue strategy and the content to be conveyed in the absence of dialogue strategy tags in the context.
The SFT dataset comprises 1000 multi-turn dialogues, with the training set and validation set being divided in an 8:2 ratio. 
We train the LoRA \cite{hu2021lora} weights of the patient simulator on the Qwen2.5-72B-Instruct \cite{yang2024qwen2} model.
% Our learning rate is set to 1e-4, and the weight decay is 0.01. We utilize bf16 precision and train for 3 epochs, consuming a total of 64 GPU hours.
Our learning rate is set to 1e-4, and we utilize DeepSpeed\footnote{\url{https://github.com/deepspeedai/DeepSpeed}} for training over 3 epochs, consuming a total of 64 GPU hours.
Finally, the weights of our patient simulator are available \textcolor{blue}{\href{https://github.com/PatientSimulator/PatientSimulator}{here}}.

% \begin{tcolorbox}[
%     colback=blue!5!white,        % 背景色
%     colframe=blue!75!white, % 边框颜色
%     coltitle=black,         % 标题字体颜色
%     fonttitle=\bfseries,    % 标题加粗
%     title=Dialogue Strategy, % 标题内容
%     sharp corners,          % 边框直角
%     enhanced,               % 启用高级特性
%     boxrule=1pt,            % 边框线宽
%     fontupper=\ttfamily,
%     width=0.5\textwidth,       % 宽度自动适应页面
% ]
% \end{tcolorbox}
\subsection{Evaluation Results}
\begin{table}[h]
\centering
\caption{Evaluation results of different patient simulators based on our defined Hallucination Rate (HR), Irrelevant Response Rate (IRR) and Anthropomorphism Score (AS). All values are presented as percentages. The final row presents the consistency results, derived from sample checks, between the performance of GPT-4o and human evaluations across these three indicators.}
\label{tab:table_1}
\begin{tabular*}{\linewidth}{@{\extracolsep{\fill}}cccc}
\toprule
\textbf{Model} & \textbf{HR}  $\downarrow$& \textbf{IRR} $\downarrow$ & \textbf{AS} $\uparrow$ \\
\midrule
Qwen2.5-72B-Instruct & 4.97 & 7.48 & 28.00\\
AgentClinic          & 3.71 & \textbf{0.93} & 31.00\\
ours                 & \textbf{0.31} & 4.79 & \textbf{87.00}\\
\midrule
Alignment with human & 99.00 & 100.00 & 90.60\\
\bottomrule
\end{tabular*}
\end{table}
% We conduct extensive experiments to evaluate the performance of our patient simulator.
% We design a set of concise and practical patient simulator metrics, which primarily include the following three indicators:
We conduct extensive experiments to evaluate the performance of our patient simulator.
We design a set of concise and practical patient simulator metrics, primarily including the following three indicators:
\begin{itemize}
    \item Hallucination Rate (HR): The proportion of dialogue turns where the patient produces responses contradicting the medical record. By inputting the medical record and each round of dialogue content, GPT-4o assigns a score (0 or 1), and the calculated proportion is evidently better when it is lower.
    \item Irrelevant Response Rate (IRR): The proportion of dialogue turns where the patient does not address the questions posed by the doctor model. It involves inputting the doctor's inquiries and the patient's responses, with GPT-4o assigning a score of 0 or 1. Since a certain level of irrelevant answers is also present in real patients, this metric does not necessarily need to be as low as possible and serves as a reference value during application.
    \item Anthropomorphism Score (AS): Analyzing the anthropomorphic behaviors exhibited by the patient agent throughout the dialogue, such as expressions of emotion, proactive questioning, and the degree of colloquialism in responses. It is scored by GPT-4o on a scale from 0 to 1, with values closer to 1 indicating a higher level of anthropomorphism.
\end{itemize}
% \begin{figure}
%     \centering
%     \includegraphics[width=\linewidth]{assets/verifier.pdf}
%     \caption{Workflow for assessing diagnostic accuracy in conversations using LLMs.}
%     \label{fig:verifier_workflow}
% \end{figure}

Our patient simulator is compared with Qwen-72B-Instruct and AgentClinic \cite{schmidgall2024agentclinic}.
% , the latter of which implements patient agents through prompt engineering on GPT-4.
AgentClinic implements patient agents through prompt engineering on GPT-4 and is used to benchmark the simulation effectiveness of our patient simulator.
Qwen-72B-Instruct benchmarks our training process.
AgentClinic utilizes multiple biased prompts that potentially interfere with HR and IRR outcomes. These biased prompts are excluded, retaining only its core system prompt.
To ensure consistency, Qwen2.5-72B-Instruct also adopts the same system prompt. 
% used in AgentClinic.

The experimental results presented in Table \ref{tab:table_1} demonstrate that our patient simulator significantly outperforms all baselines regarding Hallucination Rate.
This improvement largely stems from incorporating patient medical records into the system prompt during training. In contrast, baseline approaches depend solely on prompt engineering.
From the perspective of IRR, our method achieves a significantly lower value compared to Qwen2.5-72B-Instruct.
However, the IRR of our method is higher than that of the GPT-4-based AgentClinic.
This discrepancy may arise from differences in the underlying foundation models, as well as the selected dialogue strategy flow, where patients may ask questions proactively rather than responding to the doctor's inquiries.
It is important to note that a lower IRR is not necessarily better and should only be considered as a reference metric.
Lastly, with respect to AS, our model outperforms all baselines by a significant margin, confirming that our training paradigm is capable of successfully guiding the model to emulate a realistic dialogue strategy flow, resembling real patients.
To verify the reliability of the metrics implemented in the prompt engineering of GPT-4o, we conduct a manual random sampling inspection and calculate their consistency with human evaluations.
As indicated in the last row of Table \ref{tab:table_1}, our implementation of the three metrics demonstrates sufficient reliability.
\section{Relationship Between Inquiry and Diagnosis: Impact on Diagnostic Accuracy}

% This section outlines the experimental setup and results of exploring the relationship between inquiry and diagnosis using our patient simulator.
\subsection{Experimental Setup}
\begin{table}[h]
\centering
\caption{The distribution of the models used for inquiry and diagnosis. The o1-mini and o1-preview are used only for diagnosis due to their stronger reasoning capabilities.}
\label{tab:table_2}
\begin{tabular*}{\linewidth}{@{\extracolsep{\fill}}ccc}
\toprule
\textbf{Model}     & \textbf{Inquiry} & \textbf{Diagnosis} \\
\midrule
GPT-4o             & \textcolor{green}{\checkmark} & \textcolor{green}{\checkmark}\\
GPT-4o-mini        & \textcolor{green}{\checkmark} & \textcolor{green}{\checkmark}\\
claude-3-5-sonnet  & \textcolor{green}{\checkmark} & \textcolor{green}{\checkmark}\\
o1-mini            & \textcolor{red}{\ding{55}} & \textcolor{green}{\checkmark}\\
o1-preview         & \textcolor{red}{\ding{55}} & \textcolor{green}{\checkmark}\\
\bottomrule
\end{tabular*}
\end{table}

\begin{figure*}[t]
    \centering
    \includegraphics[width=0.95\linewidth]{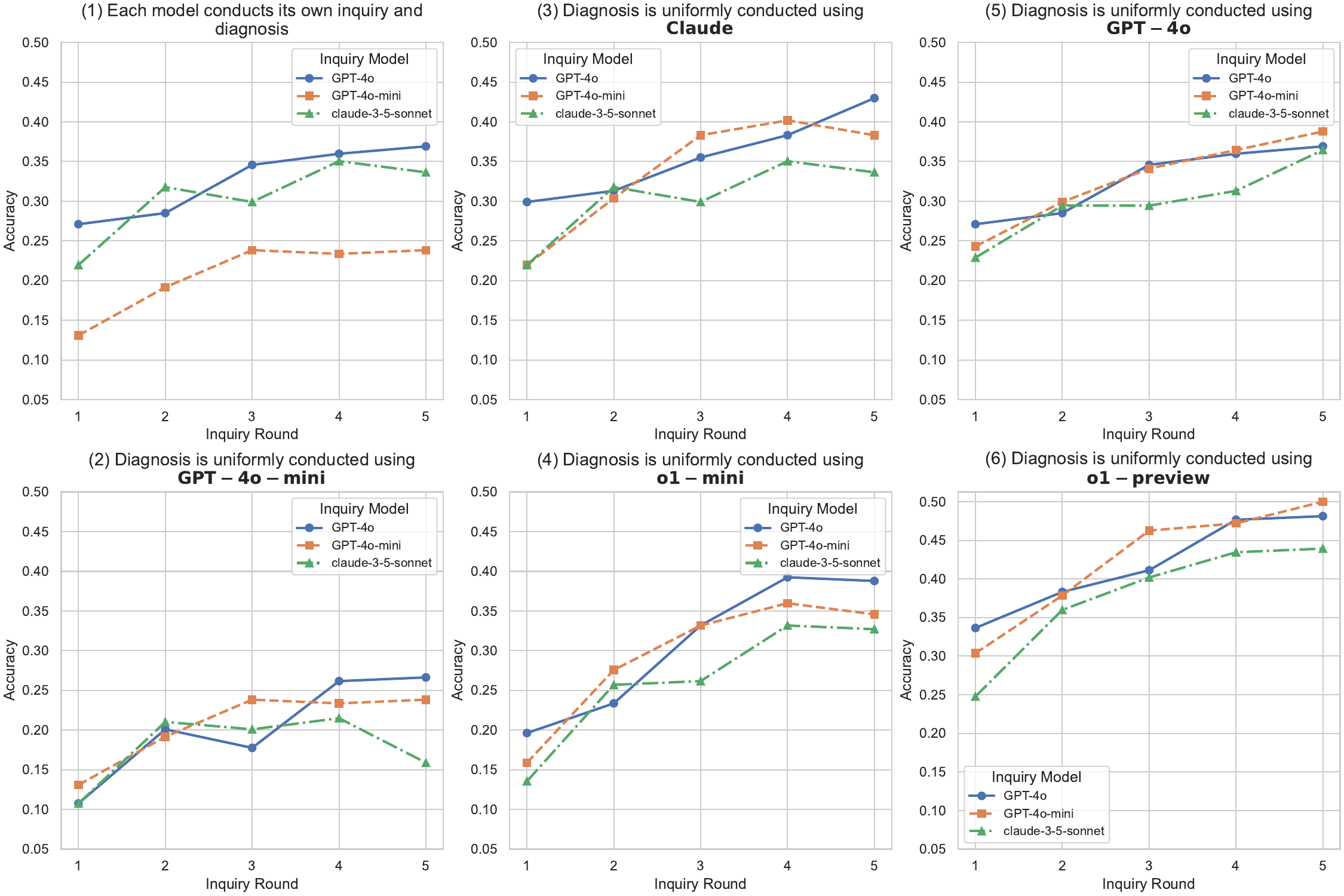}
    \caption{Patients consistently use our patient simulator, and doctors initially employ different models to interact with the simulator for fixed n rounds (x-axis, n values are 1, 2, 3, 4, 5) to generate inquiry records. These records are then diagnosed using different doctor models, and the diagnostic accuracy (y-axis) is calculated. Each experiment is conducted three times, and the average accuracy is reported.}
    \label{fig:exp1}
\end{figure*}
% Firstly, we describe the scenario of OMC.
% As depicted in Figure \ref{fig:fig1}, there are only two roles: the doctor and the patient.
% We describe the OMC scenario, involving only two roles: doctor and patient, as shown in Figure \ref{fig:fig1}.
% The process begins with the doctor collecting patient information through inquiries before providing diagnosis and medical advice.
% In practice, these inquiries often involve multiple rounds, which we denote as $n$ rounds.
% However, the number of rounds should not be excessive; generally, inquiries of up to 5 rounds are acceptable to patients.
The inquiry process typically spans $n$ rounds with various inquiry models to generate inquiry records, where $n$ is generally up to 5 rounds to ensure patient tolerance.
% The diagnosis typically occurs in the $(n+1)th$ round. , with potential updates in subsequent rounds.
% With further discussions between the doctor and patient, the diagnosis may be updated in later rounds.
We set $n$ to discrete values ranging from 1 to 5, with diagnosis made by different doctor models in the $(n+1)th$ round.
The patient side uses our patient simulator and medical records from AgentClinic's MedQA-Extend.
% Secondly, for simplicity, we set the inquiries to $n$ rounds ($1 \leq n \leq 5$) and the diagnosis to be given in the $(n+1)th$ round (i.e., the diagnosis accuracy is calculated based on the content of the $(n+1)th$ round).
% In our experiments, we conduct tests for $n$ values of 1, 2, 3, 4, and 5.
% The patient side consistently uses our patient simulator and medical records are provided by AgentClinic's MedQA-Extend.
% On the doctor's side, different inquiry models interact with the patient simulator for fixed $n$ rounds to generate inquiry records. 
% Subsequently, different doctor models were used to perform diagnosis on these inquiry records.
% Various inquiry models interact with the simulator for fixed $n$ rounds to generate inquiry records, followed by different doctor models performing diagnosis.
The specific distribution of the models used for inquiry and diagnosis is shown in Table \ref{tab:table_2}. 
To facilitate the accurate computation of diagnostic accuracy, we design a workflow (detailed in Appendix \ref{verifier_workflow_appendix}).
Tasks in the workflow are among the most common for LLMs and can yield preliminary results even without complex prompts.
% \textcolor{red}{To address the variations in the output formats of different diagnostic models}, thereby facilitating the accurate computation of diagnostic accuracy using LLMs, we design a workflow (detailed in Appendix \ref{verifier_workflow_appendix}).
% \textcolor{red}{common task}
% This process begins with inputting the complete dialogue content, followed by extracting the diagnostic results.
% These results are then subjected to necessary modifications before being compared with the ground truth (GT).
% The primary purpose of these modifications is to avoid false negatives that may arise from discrepancies such as aliases or differences in the granularity of the disease name.
% The workflow processes the complete dialogue contents to extract the diagnostic results, which are then rewritten to avoid false negatives that may arise from discrepancies.
% After that, these results are compared with the ground truth (GT).
% Tasks like result extraction, modification, and comparison are among the most common for LLMs and can yield preliminary results even without complex prompts.

% In practice, the key to achieving satisfactory outcomes lies in constructing a robust test set and conducting multiple iterations (e.g., iterations of examples and instructions).
% % 经过抽检，我们workflow与人类评估的不一致性在1%以下。
% Through sampling inspections, the inconsistency between our workflow and human evaluations remains below 1\%.
\subsection{Experimental Results}
% 我们的实验结果如Figure 5所示。
Our experimental results are presented in Figure \ref{fig:exp1}.
Patients consistently utilize our patient simulator, while doctors interact with the simulator using various models for a fixed number of rounds (x-axis, where n values are 1, 2, 3, 4, 5) to generate inquiry records. 
Subsequently, these records are diagnosed by five different doctor models, as shown in Table \ref{tab:table_2}, and the diagnostic accuracy (y-axis) is computed.

Firstly, we analyze Subfigures 2 to 6, excluding the first Subfigure in the upper left corner of Figure \ref{fig:exp1}.
These five Subfigures present the accuracy rates of the same three sets of inquiries processed through five different diagnostic models.
By examining each Subfigure individually, it becomes apparent that under the same inquiry rounds and diagnostic models, there are significant differences in the accuracy rates of inquiries generated by different models.
For example, in Subfigure 6, after 5 inquiry rounds and under the o1-preview diagnostic model, the accuracies for Claude, GPT-4o and GPT-4o-mini \cite{openai4o} are 0.439, 0.481, and 0.5, respectively.
Furthermore, across all five Subfigures, the inquiries produced by the model claude-3-5-sonnet consistently exhibit relatively lower accuracy levels, regardless of the diagnostic model used.
These indicate that \textbf{there are significant differences in inquiry capabilities among the different models}.
% Additionally, by comparing the accuracy across the five Subfigures for the same inquiry iterations, it becomes evident that diagnostic capabilities vary among the models.
% o1-preview demonstrates the strongest diagnostic capability, whereas GPT-4o-mini shows the weakest performance.

Secondly, by comparing the accuracy rates of the same inquiry rounds and models in Subfigures 2 to 6, we observe that different models exhibit varying diagnostic capabilities.
Among them, o1-preview demonstrates the strongest diagnostic ability, while GPT-4o-mini shows the weakest.
This result correlates with the inherent reasoning capabilities of the models, aligning with intuitive expectations.
By further integrating the performance of diagnostic and inquiry abilities, it is observed that there is no significant correlation between the two.
For instance, while GPT-4o-mini exhibits weaker diagnostic capabilities, it performs relatively well in inquiry tasks, whereas GPT-4o demonstrates strong performance in both areas.
This observation suggests that when developing medical AI models, if a single model struggles to excel in both inquiry and diagnostic abilities, dividing the tasks into two specialized models could serve as a viable solution.

Thirdly, comparing Subfigure 2 with Subfigures 3 to 6 for the same inquiry rounds and models reveals that the accuracy rates in Subfigure 2 are significantly lower than those in Subfigures 3 to 6.
This is due to the weaker diagnostic capability of GPT-4o-mini, leading to a lower ceiling for the final accuracy.
Conversely, comparing Subfigure 6 with the others for the same rounds and inquiry models shows that Subfigure 6 surpasses the others in accuracy rates.
This is attributed to the superior diagnostic ability of o1-preview, resulting in a higher ceiling.
Observations from Subfigures 1 and 3 to 6 indicate that diagnostic accuracy increases significantly with more inquiry rounds.
Furthermore, regardless of the diagnostic model used, records based on Claude inquiries consistently perform poorly.
Hence, we conclude that \textbf{inquiry and diagnosis adhere to the Liebig's law}: if the quality of the inquiry is insufficient, achieving good results is challenging even with strong diagnostic capabilities, and vice versa.
% GPT-4o、GPT-4o-mini和claude-3-5-sonnet分别作为询问模型，按照询问轮次拆分，在4种询问类型上的分布对比。横轴为询问模型，纵轴为4种询问类型的占比。
\section{Inquiry Differences Among Models}
% 为了分析不同询问模型的差异在哪儿，我们进一步分析了GPT-4o、GPT-4o-mini和claude-3-5-sonnet的5轮询问记录。
% 结合已有的对话策略标签和医生询问内容，我们将医生询问分成4类：(1) main complaint inquiry; (2) specification of known symptoms; (3) inquiry about accompanying symptoms; (4) gathering family or medical history.
% To better analyze the differences among various inquiry models, we further examine the inquiry records of GPT-4o, GPT-4o-mini, and Claude-3-5-sonnet across five rounds.

% 4种询问类型举例

\subsection{Four Types of Inquiry}

\begin{figure}
    \centering
    \includegraphics[width=\linewidth]{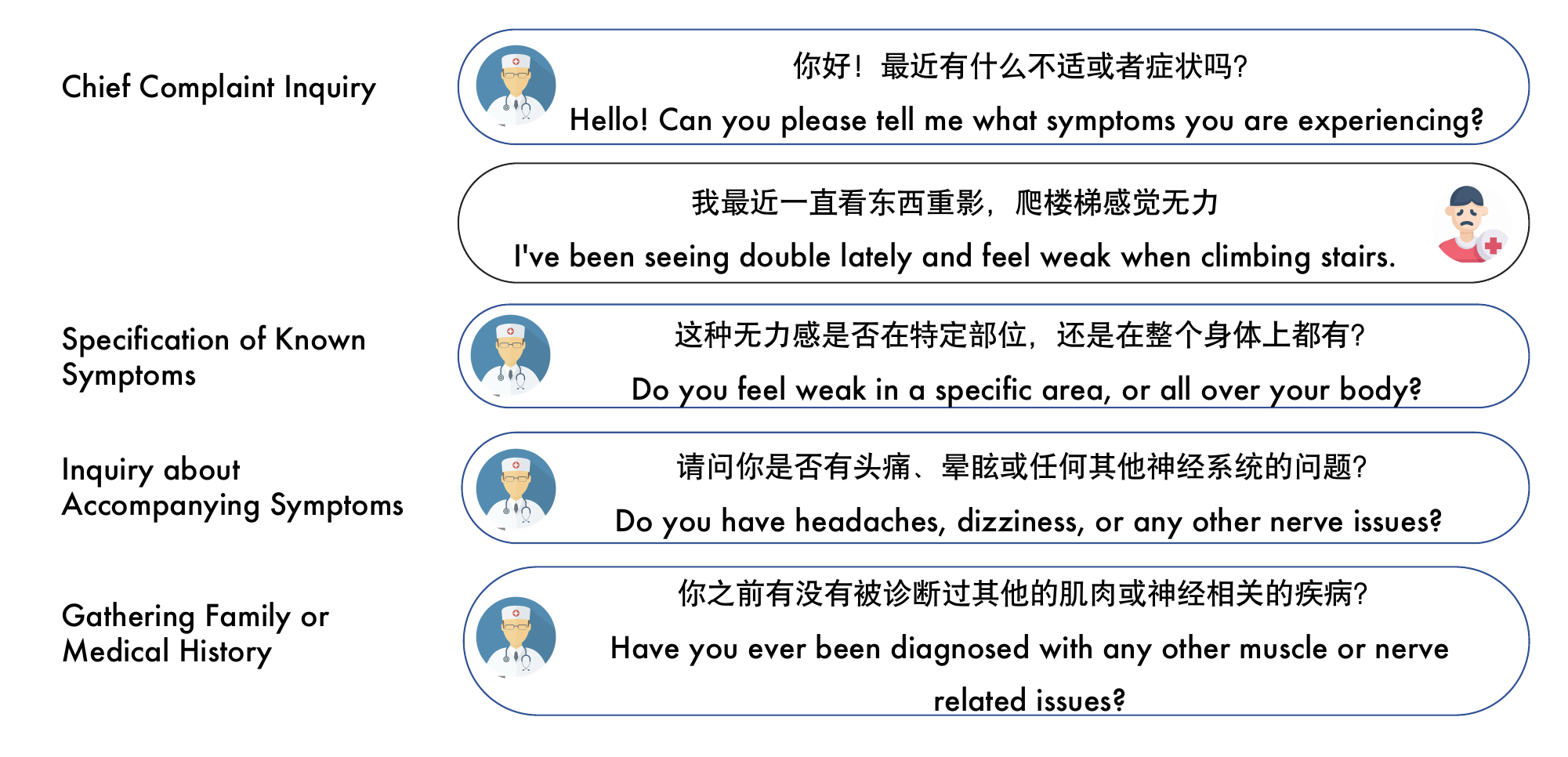}
    \caption{Examples of four types of inquiry with D representing the doctor and P representing the patient in the figure.}
    \label{fig:inquiry_type_example}
\end{figure}

Based on the examples in our inquiry records and systematic descriptions in relevant medical materials \cite{trousseau1873lectures, adler1997history,bickley2012bates,swartz2014textbook}, we categorize the doctors' inquiries into four types: (1) chief complaint inquiry; (2) specification of known symptoms; (3) inquiry about accompanying symptoms; (4) gathering family or medical history, as shown in Figure \ref{fig:inquiry_type_example}.
% Examples of the four types are provided in Figure \ref{fig:inquiry_type_example}.
% 结合询问记录中例子（cases）和相关医学书籍中的系统性描述，
A detailed discussion is in Appendix \ref{appendix_Discussion}.
\subsection{Experimental Results}

\begin{figure*}[t]
    \centering
    \includegraphics[width=0.95\linewidth]{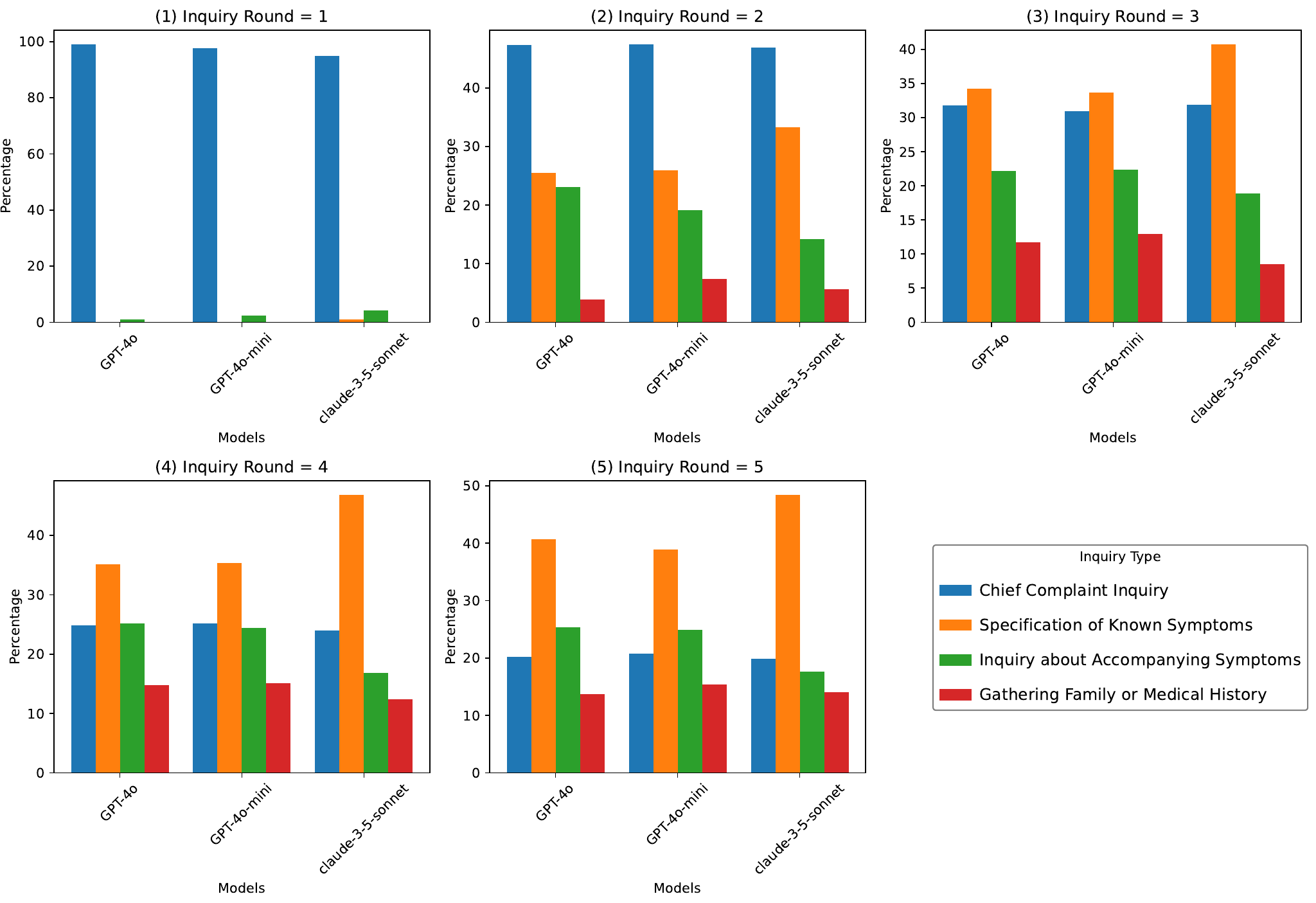}
    \caption{The comparison focuses on the distribution of four inquiry types across GPT-4o, GPT-4o-mini, and Claude-3-5-sonnet as inquiry models, segmented by inquiry rounds. The x-axis represents the inquiry models, while the y-axis indicates the proportion of the four inquiry types.}
    \label{fig:exp2}
\end{figure*}
% 我们采用GPT-4o对上述3份询问记录进行了打标，使用到的prompt参见附录B。
We employ GPT-4o to annotate the inquiry records into above four types, with the prompt used detailed in the Appendix \ref{appendix_b}.
Our experimental results are shown in Figure \ref{fig:exp2}.
% Segmented by inquiry rounds, the comparison focuses on the distribution of four inquiry models.
% The x-axis represents the inquiry models, while the y-axis indicates the proportion of the four inquiry types.

Firstly, as shown in Subfigure 1 of Figure \ref{fig:exp2}, in the vast majority of cases, all inquiry models choose to ask about the chief complaint during the first round. This aligns with expectations, as the doctor models do not possess any information about the patient during the initial round and thus typically begin with a question such as "What symptoms have you been experiencing that brought you here today?".
However, there is a small subset involving inquiries about accompanying symptoms, particularly with the use of GPT-4o-mini and Claude. 
Such initial questions often include: "Hello, could you tell me if you've had any discomfort in recent days, \textbf{like fever, cough, or any other uneasy feelings}?" or "Hello, \textbf{you seem a bit pale}; have you been experiencing any symptoms \textbf{like dizziness, fatigue, or loss of appetite}?".
Although whether these instances should be tagged as inquiries about accompanying symptoms remains debatable, the comparison shows that these inquiries indeed interfere with the collection of the patient's chief complaint. This might be the main reason why GPT-4o consistently performs the best in the first round in subfigures 2 to 6 of Figure \ref{fig:exp1}.

Secondly, as shown in Subfigures 2–5 of Figure \ref{fig:exp2}, Claude demonstrates a significantly higher proportion of specification of known symptoms during multi-turn inquiry compared to other models.
This leads to a noticeable reduction in the proportions of other inquiry types.
Considering that each type of inquiry is crucial for the diagnostic process, we hypothesize that this might indicate a relative weakness in Claude's overall inquiry capability compared to other models.
Correspondingly, in Subfigures 2–6 of Figure \ref{fig:exp1}, the inquiry records generated by Claude are generally associated with the lowest final diagnostic accuracy.
Furthermore, when comparing GPT-4o and GPT-4o-mini, the latter consistently exhibits a higher proportion of gathering family or medical history across turns (except for the fourth turn).
Based on Subfigure 6 of Figure \ref{fig:exp1} (where o1-preview is used as the diagnostic model), the contribution of family history to diagnostic accuracy becomes evident starting from the third turn.
The focus on subfigure 6 is motivated by the fact that o1-preview demonstrates the strongest diagnostic capability among all models, allowing us to minimize the confounding effects of different levels of diagnostic performance.
\section{Related Works}

\subsection{Large Language Models in Medicine}
Large language models (LLMs) in medicine are categorized into two types: general-purpose LLMs and medical-specific models.
General-purpose LLMs are further divided into open-source and closed-source categories.
Examples of open-source models include LLaMA \cite{dubey2024llama}, Qwen \cite{yang2024qwen2}, Mixtral \cite{jiang2024mixtral}, and DeepSeek \cite{liu2024deepseek}, while closed-source models include GPT-4o \cite{openai4o}, o1-preview \cite{openaio1}, Claude \cite{claude} and Gemini \cite{team2023gemini}.
The primary goal of optimizing a general-purpose LLMs is to enhance its broad applicability, ensuring strong performance across a variety of tasks, including, naturally, medical tasks.
Building on the core strengths of general-purpose LLMs, numerous researchers focus on developing specialized models for the medical domain.
These models enhance their performance in the medical field through prompt engineering, continual pre-training, supervised fine-tuning (SFT), and reinforcement techniques. \cite{tian2023chimed,saab2024capabilities,chen2024huatuogpt,zhang2024ultramedical,singhal2025toward}

\subsection{The Evaluation of Language Models in Medicine}
The benchmarks to evaluate LLMs in medicine 
% studies validate the performance of LLMs using various benchmarks, demonstrating their significant potential for future medical practice.
can be categorized into static and dynamic types based on whether they provide a simulated environment.

Static benchmarks primarily assess medical knowledge and typically use a multiple-choice format.
The MedQA \cite{jin2021disease} dataset contains question-answer pairs derived from the US, Mainland China, and Taiwan Medical Licensing Exams. It features 4-5 multiple-choice questions with correct answers. 
% , supported by explanations or references.
% These questions, ranging from diagnosis to treatment selection, are often challenging even for medical students.
The LLMs receive comprehensive context, including patient history, demographics, and symptoms, to generate responses.
And similar multiple-choice formats are employed by PubMedQA \cite{jin2019pubmedqa}, MedMCQA \cite{pal2022medmcqa}, MMLU clinical topics \cite{hendrycks2020measuring}, and MultiMedQA \cite{singhal2023large}.

Dynamic benchmarks assess the performance of doctor models through role-playing scenarios involving doctors and patients, utilizing LLMs.
AMIE \cite{tu2024towards} diagnoses simulated patients through history-taking. 
AgentClinic \cite{schmidgall2024agentclinic} is an open-source multimodal benchmark designed to assess the capability of LLMs to function as agents in simulated clinical settings.
% Compared to static benchmarks, numerous studies  demonstrate the potential of dynamic methods in simulating real clinical scenarios, offering valuable insights for the development of better AI doctors.
Additionally, many other studies \cite{li2024mediq,qiu2024interactive,li2024agent,johri2023guidelines,tang2023medagents} provide simulated clinical environments to evaluate or enhance physician models.
However, in these studies, patient simulations predominantly rely on prompt engineering, which does not accurately replicate real patient behavior.
% Unlike passive simulated responses, real patients often express anxiety about their conditions, ask questions to alleviate concerns, and may not always be cooperative.
% They might terminate the conversation or refuse to answer if doctors repeatedly question them.
% These complexities cannot be adequately addressed through prompt engineering alone, highlighting the need for innovative methods in patient simulation.
Furthermore, the relationship between inquiry and diagnosis remains unexplored.

\section{Conclusion}

In this paper, we use real doctor-patient dialogue strategies to guide the training of our patient simulator, resulting in a simulation that has significantly fewer hallucinations and more accurately resembles a real patient. Utilizing this simulator for comprehensive experiments uncovers significant differences in inquiry strategies across various models and demonstrates that inquiry and diagnosis adhere to Liebig's law. We classify inquiries into four categories based on data cases and diagnostic definitions.% : (1) chief complaint inquiry; (2) specification of known symptoms; (3) inquiry about accompanying symptoms; and (4) gathering family or medical history. 
We label and analyze the distribution of inquiries across four types, identifying specific differences in inquiry strategies based on variations in distribution and diagnostic accuracy. 
% Our results suggest that optimizing the allocation of limited inquiry opportunities (commonly 3 to 5 rounds are acceptable to patients) is a significant area for further research.
Our results suggest that optimizing the allocation of inquiries within typically 3 to 5 rounds, which are acceptable to patients, presents a valuable research problem.
% In this paper, we extract real dialogue strategy flows from authentic doctor-patient conversations, and after manual selection, these flows are combined with patient records to generate synthetic data.
% These dialogue strategy flows, along with patient records, guide the training of a patient simulator, resulting in a patient simulation with significantly fewer hallucinations, and more closely resembling a real patient.
% Based on this simulator, we conduct extensive experiments to explore the relationship between inquiries and diagnoses in medical consultations, as well as their impact on the final diagnostic accuracy.
% Our experimental results demonstrate: 1) There are significant differences in inquiry strategies among different models; 2) Inquiries and diagnoses adhere to the Liebig's law.
% Furthermore, we classify inquiries into four types based on data cases and relevant definitions in diagnostics: (1) chief complaint inquiry; (2) specification of known symptoms; (3) inquiry about accompanying symptoms; and (4) gathering family or medical history. 
% We label and analyze the distribution of inquiries generated by different models according to these four types. 
% By comparing distribution differences and diagnostic accuracy variations, we uncover the specific differences in inquiry strategies among the models.
% Our results indicate that how to allocate limited inquiry opportunities (commonly 3 to 5 rounds are acceptable to patients) is a research-worthy issue, which we intend to explore further in future work.

\section*{Limitations}
Due to the lack of multimodal information in open-source doctor-patient dialogue data, our patient simulator does not support sending images or videos. 
During the selection of dialogues, we retained only initial consultations for simplification, which limits the ability of our patient simulator to effectively simulate follow-up consultations. 
Additionally, since the available open-source medical records data are in Chinese, we chose to synthesize only Chinese dialogues, which constrains our simulator's performance in English dialogue scenarios.
Finally, this paper does not propose a specific approach to allocate questions within the limited rounds of the inquiry stage, leaving this aspect for future work.
\bibliography{custom}

\appendix

% \section{Example Appendix}
% \label{sec:appendix}

\section{Candidate Set of Dialogue Strategy Tags}
\label{candidate_set_of_dialogue_strategy_tags}

% \begin{table*}
\resizebox{\linewidth}{!}{
\begin{tabular}{c|l}
% \centering
\hline
\textbf{Doctor Dialogue Strategy Labels} & \textbf{Description} \\
\hline
{[Greeting]} & The doctor initiates the conversation by greeting the patient. \\
{[Explanation]} & The doctor explains the patient's condition, treatment plan, or medication use. \\
{[Answering]} & The doctor responds to the patient's questions or concerns. \\
{[Clarification]} & The doctor or patient clarifies certain issues. \\
{[Medical Advice]} & The doctor offers health advice or lifestyle guidance. \\
{[Confirmation]} & The doctor or patient confirms certain information or understanding. \\
{[Concern]} & The doctor expresses concern and attention for the patient. \\
{[Comfort]} & The doctor shows care and comfort to the patient. \\
{[Diagnosis]} & The doctor identifies the patient's condition based on symptoms and examination.\\
{[Education]} & The doctor identifies the illness or other problem of the patient. \\
{[Chief Complaint Inquiry]} & The doctor asks the patient to describe their primary health concern.\\
{[Recommendation]} & The doctor gives health advice or suggests lifestyle changes. \\
{[Inquiring about Symptoms]} & The doctor asks about the patient's symptoms, medical history, and related information. \\
{[Inquiry about Accompanying Symptoms]} & The doctor asks about other symptoms alongside the main issue.\\
{[Gathering Family or Medical History]} & The doctor asks about the patient's past medical history or family medical history. \\
{[Evaluation]} & The doctor assesses the patient's symptoms. \\
{[Arrangement]} & The doctor arranges for follow-up tests or appointments. \\
{[Prescription]} & The doctor prescribes medication or treatment plans for the patient. \\
{[Farewell]} & The doctor concludes the conversation. \\
\hline
\end{tabular}
% \end{table*}
}

% \vspace{1cm} % 可根据需要调整间隔

\resizebox{\linewidth}{!}{
% \begin{table*}
\begin{tabular}{c|l}
\hline
\textbf{Patient Dialogue Strategy Labels} & \textbf{Description} \\
\hline
{[Greeting]} & The patient initiates the conversation by greeting the doctor. \\
{[Describe Condition]} & The patient describes their symptoms or medical history. \\
{[Detail Symptoms]} & The patient elaborates on specific physical discomforts or symptoms. \\
{[Ask Questions]} & The patient asks further questions regarding the doctor's advice or the condition. \\
{[Confirm]} & The patient confirms or shows understanding of what the doctor has said. \\
{[Express Concerns]} & The patient expresses worries about the condition or the treatment outcome. \\
{[Seek Help]} & The patient requests support or assistance from the doctor. \\
{[Provide Information]} & The patient proactively offers relevant health information or past medical history. \\
{[Discuss Treatment Options]} & The patient discusses possible treatment options with the doctor. \\
{[Disagree]} & The patient expresses a different opinion on the doctor's advice or diagnosis. \\
{[Explanation Request]} & The patient asks the doctor to further explain the test results or treatment plan. \\
{[Seek Advice]} & The patient requests professional advice or suggestions from the doctor. \\
{[Complaint or Feedback]} & The patient offers opinions or suggestions regarding the medical service or treatment process. \\
{[Request Prescription]} & The patient asks the doctor to prescribe medication. \\
{[Inquire about Treatment Options]} & The patient asks about feasible treatment options and expected outcomes. \\
{[Share Feelings]} & The patient shares their feelings about the condition or treatment, such as pain, anxiety, etc. \\
{[Request Recommendation]} & The patient asks the doctor to recommend other specialists or tests. \\
{[Thanks]} & The patient expresses gratitude for the doctor's help or advice. \\
{[Disagree]} & The patient expresses a different opinion on the doctor's advice or treatment plan. \\
{[Emotional Expression]} & The patient expresses emotional reactions to the condition, such as depression, anger, gratitude, etc. \\
{[Express Concerns]} & The patient expresses anxiety or concerns about their health condition or treatment plan. \\
{[Ask about Side Effects]} & The patient inquires about possible side effects of the medication or treatment. \\
{[Seek Understanding]} & The patient hopes the doctor will provide more explanation and understanding of their condition. \\
{[Ask about Follow-up Arrangements]} & The patient inquires about subsequent tests, follow-up visits, or treatment plans. \\
{[Stop]} & The patient ends the conversation. \\
\hline
\end{tabular}
% \end{table*}
}

\section{Dialogue Strategy Flows}
\begin{figure}[htbp]
    \centering
    \includegraphics[width=\linewidth]{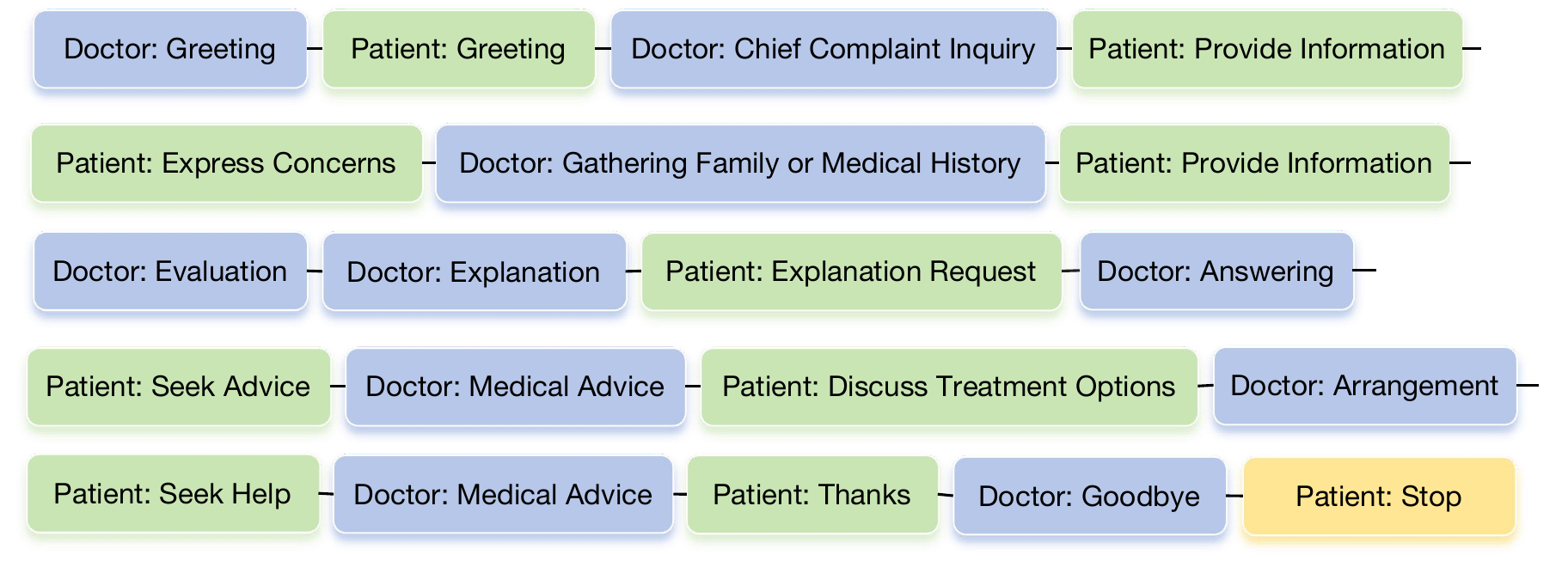}
    \caption{Example for a dialogue strategy flow.}
    \label{fig:diag_strategy_flow}
\end{figure}

This appendix presents examples of high-quality dialogue strategy flows manually selected from the deduplicated set. Each dialogue’s tags are concatenated sequentially to form a structured dialogue strategy flow (see Figure \ref{fig:diag_strategy_flow}).
\label{examples_strategy_appendix}

\section{System Prompt of Patient Simulator}
\begin{figure}[htbp]
    \centering
    \includegraphics[width=\linewidth]{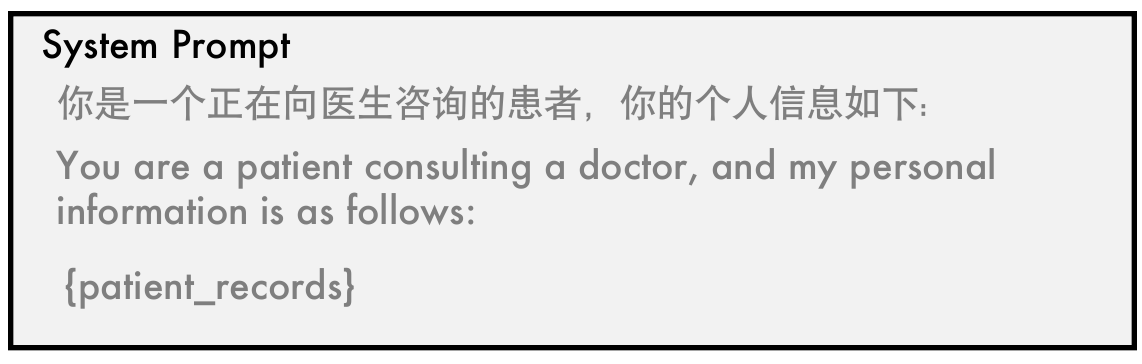}
    \caption{The system prompt of our patient simulator.}
    \label{fig:patient_sys}
\end{figure}

The detailed system prompt of our patient simulator is shown in Figure \ref{fig:patient_sys}.
\label{patient_sys_appendix}

\section{The Discussion of Inquiry Type}
\label{appendix_Discussion}
The detailed discussion of four inquiry types is presented below.

\textbf{Chief Complaint Inquiry}: This refers to asking patients about their most significant discomfort, the most prominent symptoms, or signs they experience, which often represent the primary reason for the visit. A precise chief complaint provides an initial indication of the severity and urgency of the condition and offers diagnostic clues for identifying potential systemic diseases.

\textbf{Specification of Known Symptoms}
% generalized abdominal pain typically suggests widespread pathology or peritoneal involvement
% leg edema for 4 days
% \begin{itemize}

\underline{Onset and duration of illness}: Each disease has unique characteristics regarding its onset and progression. Thus, detailed inquiry into the onset of illness is essential for differential diagnosis. Some diseases have an acute onset, such as cerebral embolism, while others progress more slowly, like pulmonary tuberculosis. The duration of illness refers to the time from disease onset to the point of clinical consultation or hospitalization. If multiple symptoms appear, it is essential to trace back to the time of the initial symptom and document the entire medical history in chronological order. For instance, the patient may experience palpitations for 3 months and recurrent nocturnal dyspnea for 2 weeks.

\underline{The characteristics of the main symptoms}: The location, nature, duration, and intensity of symptoms, along with factors that alleviate or worsen them, are essential for diagnosing the affected system or organ and determining the pathological changes' site, extent, and nature. For instance, upper abdominal pain often points to issues with the stomach, duodenum, or pancreas, while acute pain in the right lower abdomen typically suggests appendicitis. The type of pain—whether burning, colicky, distention, or dull—and whether symptoms are continuous or intermittent, as well as their onset and relief patterns, are diagnostically significant.
% \end{itemize}

\textbf{Inquiry about Accompanying Symptoms}: On the basis of the primary symptoms, a series of accompanying symptoms often emerge. These accompanying symptoms are crucial for differential diagnosis or indicating possible complications. For instance, diarrhea may be a common symptom of various underlying causes, making it difficult to diagnose a specific disease based solely on this symptom. However, by inquiring about the accompanying symptoms, the diagnostic direction becomes clearer. For example, diarrhea accompanied by vomiting may suggest acute gastroenteritis caused by consumption of contaminated food or toxic substances, whereas diarrhea with a sensation of incomplete evacuation, when considered along with seasonality and dietary habits, is more likely associated with dysentery.

\textbf{Gathering Family or Medical History} 
% \begin{compactitem}

\underline{Family history}: It is important to inquire about the health and disease conditions of the patient's parents, siblings, and children. Particular attention should be paid to whether there are diseases similar to that of the patient, or hereditary diseases such as hemophilia, albinism, familial hypothyroidism, diabetes, and mental illnesses.

\underline{Diagnosis and treatment history}: If the patient has already received medical treatment at other healthcare facilities prior to this visit, it is essential to inquire about the previous diagnoses, treatments, and their outcomes. If treatment has been administered, a thorough understanding of the medications used, including their names, dosages, durations, and effects, is necessary to inform the current diagnosis and treatment plan.

\underline{Past medical history (PMH)}: PMH encompasses the patient's prior health status and previously diagnosed conditions, including infectious diseases, injuries, surgical procedures, immunization records, and allergy history, with particular emphasis on factors closely related to the current illness.
% \end{compactitem}

\section{Inquiry Type Annotation Prompt}
\begin{figure}[htbp]
    \centering
    \includegraphics[width=\linewidth]{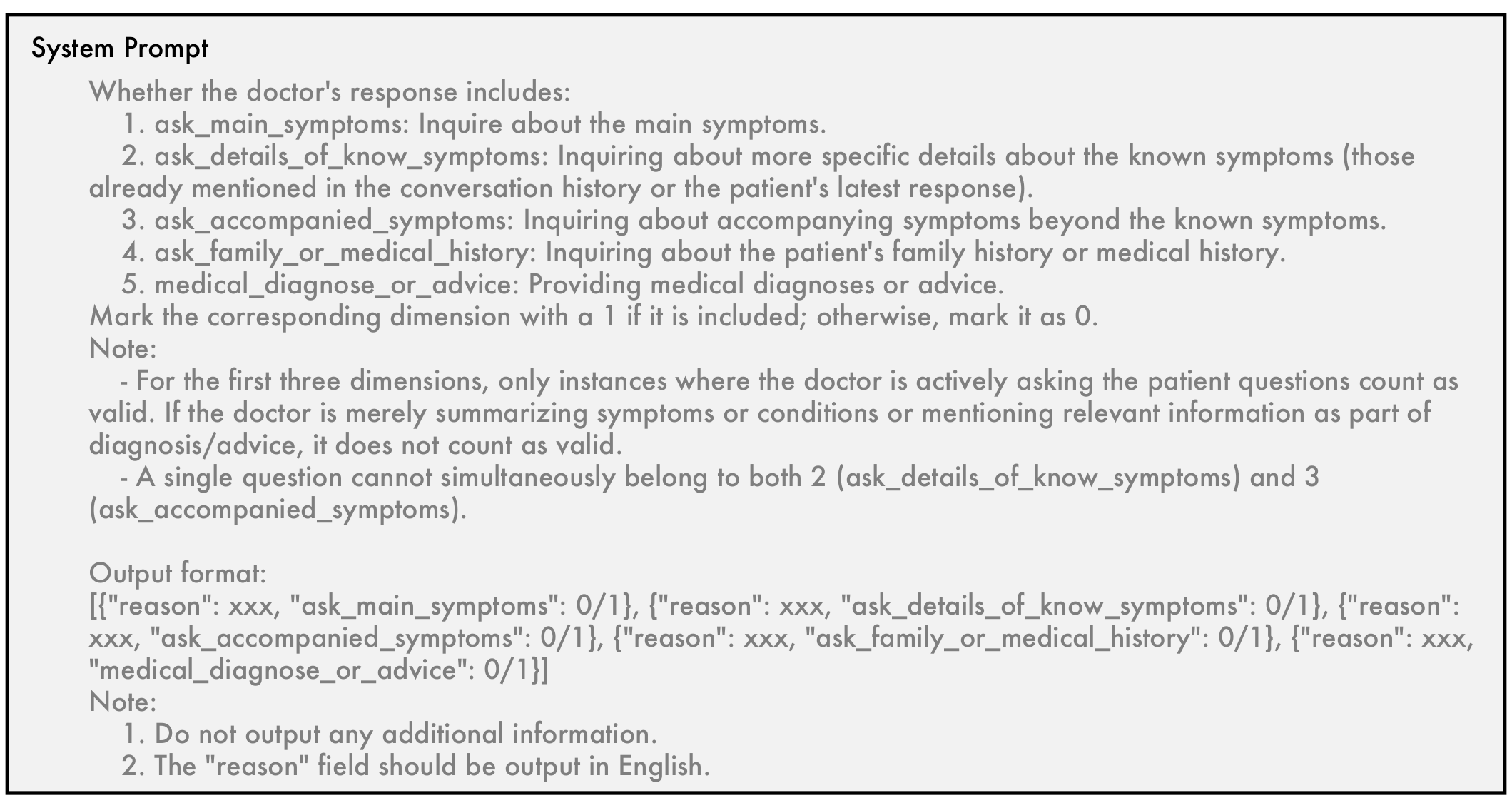}
    \caption{Inquiry type annotation prompt.}
    \label{fig:inquiry_type_annotation_prompt}
\end{figure}

The detailed prompt of inquiry type annotation is shown in Figure \ref{fig:inquiry_type_annotation_prompt}.
\label{appendix_b}

\section{Workflow for assessing diagnostic accuracy}
\label{verifier_workflow_appendix}
\begin{figure}[htbp]
    \centering
    \includegraphics[width=\linewidth]{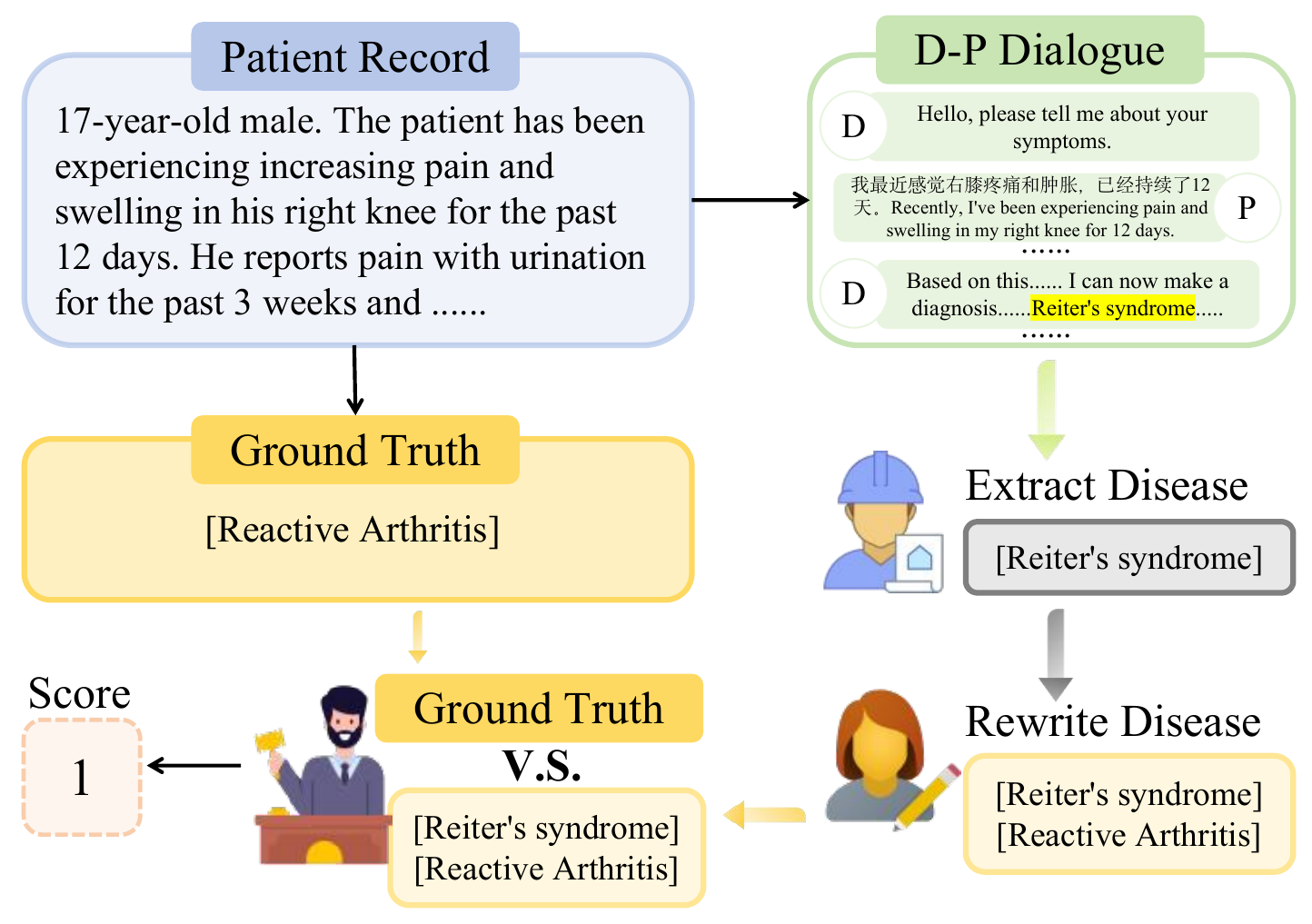}
    \caption{Workflow for assessing diagnostic accuracy in conversations using LLMs.}
    \label{fig:verifier_workflow}
\end{figure}

To address the variations in the output formats of different diagnostic models and calculate accurate diagnostic accuracy computation using LLMs, we designed a standardized workflow. This workflow is illustrated in Figure \ref{fig:verifier_workflow}. 

The workflow processes the complete dialogue contents to extract the diagnostic results, which are then rewritten to avoid false negatives that may arise from discrepancies.
After that, these results are compared with the ground truth (GT).
Tasks mentioned above, like result extraction, are among the most common for LLMs and can yield preliminary results even without complex prompts.
In practice, the key to achieving satisfactory outcomes lies in constructing a robust test set and conducting multiple iterations (e.g., iterations of examples and instructions).
Through sampling inspections, the inconsistency between our workflow and human evaluations remains below 1\%.

\end{document}